\documentclass{article}
\PassOptionsToPackage{numbers, compress}{natbib}
\usepackage[final]{neurips_2022}

\usepackage[utf8]{inputenc} 
\usepackage[T1]{fontenc}    
\usepackage{hyperref}       
\usepackage{url}            
\usepackage{booktabs}       
\usepackage{amsfonts}       
\usepackage{nicefrac}       
\usepackage{microtype}      
\usepackage{xcolor}         

\usepackage{amsmath}
\usepackage{amsthm}
\usepackage{enumitem}
\usepackage{multirow}
\usepackage{xspace}
\usepackage{graphicx}

\newcommand{\method}{PaxNet\xspace}

\newtheorem{definition}{Definition}

\DeclareMathOperator*{\concat}{\scalebox{1}[1.0]{$\parallel$}}
\newcommand{\angstrom}{\text{\normalfont\AA}}

\title{Physics-aware Graph Neural Network for Accurate RNA 3D Structure Prediction}

\author{%
  Shuo Zhang$^{1,2}$, Yang Liu$^{2}$, Lei Xie$^{1,2,3}$ \\
  $^1$Ph.D. Program in Computer Science, Graduate Center, City University of New York \\
  $^2$Department of Computer Science, Hunter College, City University of New York \\
  $^3$Helen \& Robert Appel Alzheimer’s Disease Research Institute, \\
  Feil Family Brain \& Mind Research Institute, \\
  Weill Cornell Medicine, Cornell University \\
\texttt{szhang4@gradcenter.cuny.edu, thomasliuy@gmail.com, lxie@iscb.org} \\
}

\begin{document}

\maketitle

\begin{abstract}
  Biological functions of RNAs are determined by their three-dimensional (3D) structures. Thus, given the limited number of experimentally determined RNA structures, the prediction of RNA structures will facilitate elucidating RNA functions and RNA-targeted drug discovery, but remains a challenging task. In this work, we propose a Graph Neural Network (GNN)-based scoring function trained only with the atomic types and coordinates on limited solved RNA 3D structures for distinguishing accurate structural models. The proposed Physics-aware Multiplex Graph Neural Network (\method) separately models the local and non-local interactions inspired by molecular mechanics. Furthermore, \method contains an attention-based fusion module that learns the individual contribution of each interaction type for the final prediction. We rigorously evaluate the performance of \method on two benchmarks and compare it with several state-of-the-art baselines. The results show that \method significantly outperforms all the baselines overall, and demonstrate the potential of \method for improving the 3D structure modeling of RNA and other macromolecules. Our code is available at \url{https://github.com/zetayue/Physics-aware-Multiplex-GNN}.
\end{abstract}

\section{Introduction}
RNAs play an essential role in many cellular processes in organisms, such as mediating gene translation, regulating gene expression, catalyzing biochemical reactions, and transferring cellular information~\cite{yao2019cellular,dethoff2012functional,warner2018principles}. These cellular functions are determined by their unique three-dimensional (3D) structures. However, due to the high cost of experimental methods, the high-resolution RNA 3D structures deposited in the Protein Data Bank (PDB) are still very scarce~\cite{burley2021rcsb}. Thus, there is an urgent need to develop computational RNA 3D structure prediction methods for elucidating RNA functions and promoting RNA-targeted drug discovery~\cite{kulkarni2021current}.

At the current state, most existing computational methods for the prediction of RNA 3D structures are non-deep learning methods, and are based on physics or knowledge-guided approaches like coarse-grained representations~\cite{boniecki2016simrna,jonikas2009coarse,zhang2018isrna}, fragment-assembling~\cite{cao2011physics,popenda2012automated,zhao2012automated}, statistical potentials~\cite{bernauer2011fully,capriotti2011all,wang20153drnascore,zhang2020all}, etc. Only a limited number of deep learning approaches have
been proposed to be scoring functions for evaluating RNA 3D structural models~\cite{li2018rna3dcnn,townshend2021geometric}. Specifically, ARES~\cite{townshend2021geometric} is a Graph Neural Network (GNN) that outperforms the previous state-of-the-art methods using only a small number of RNA structures for training without any assumptions about structural characteristics. This has demonstrated the power of deep learning on tackling the challenge of RNA 3D structure prediction. 

In this work, we aim at developing a new deep learning-based approach to predict RNA 3D structures. Inspired by the physical principles in molecular mechanics methods~\cite{schlick2010molecular} that separately considers the local and non-local interactions when computing molecular energy, we propose \underline{P}hysics-\underline{a}ware Multiple\underline{x} Graph Neural \underline{Net}work (\method), which is a GNN based on message passing scheme~\cite{gilmer2017neural}. \method represents each RNA 3D structure using a two-plex (or layer) multiplex graph, where one plex only contains local interactions, and the other plex further contains non-local interactions. \method takes the multiplex graphs as input and uses different message passing modules to incorporate the geometric information in different kinds of interactions. Furthermore, with a fusion module, the contribution of each kind of interaction can be learned and fused to compute the final prediction. Similar to ARES, \method only uses atomic types and coordinates as input without having any assumptions about other structural and physiochemical characteristics.

To verify the effectiveness of \method, we train and evaluate \method on the datasets used in~\cite{townshend2021geometric} for the RNA 3D structure prediction. Our \method outperforms previous state-of-the-art methods on all benchmarks in general. The results suggest that \method provides a powerful method for the representation learning of RNA 3D structures. Our main contributions are as follows:
\begin{itemize}[leftmargin=10pt]
\item We propose a novel GNN, Physics-aware Multiplex Graph Neural Network (\method) for the prediction of RNA 3D structures. Inspired by molecular mechanics, \method models the local and non-local interactions differently by building multiplex graphs to represent RNA structures and designing diverse message passing modules.
\item Comprehensive experiments for RNA 3D structure prediction are conducted to demonstrate that our proposed \method outperforms the state-of-the-art methods.
\end{itemize}

\section{Method}

\begin{figure*}[t]
    \centering
	\includegraphics[width=0.9\columnwidth]{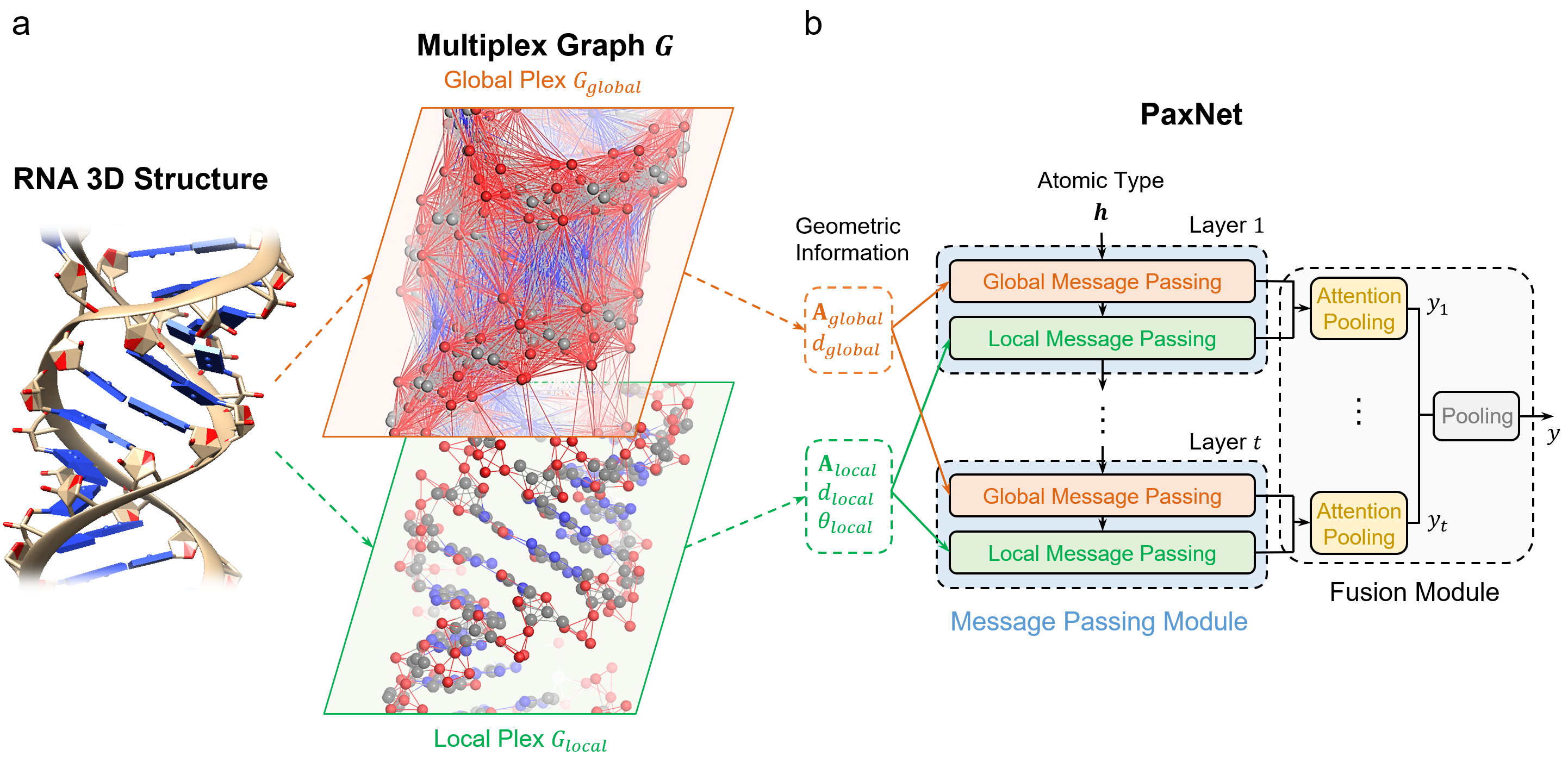}
	\vskip -0.05in
	\caption{\label{fig:method}\textbf{Illustration of \method pipeline.} (a) Construction of multiplex graph $G = \{G_{global}, G_{local}\}$ based on a given RNA 3D structure. (b) Overall architecture of \method, which consists of message passing modules and a fusion module.}
	\vskip -0.1in
\end{figure*}

The overview of Physics-aware Multiplex Graph Neural Network (\method) is shown in Figure~\ref{fig:method}. Given a RNA 3D structure, a corresponding multiplex graph $G$ (Figure~\ref{fig:method}a) is built, where each plex of $G$ contains a different group of atomic interactions. Then \method (Figure~\ref{fig:method}b) uses different message passing modules to update the atomic type embeddings $\boldsymbol{h}$ with the geometric information in different plex of $G$ accordingly. To learn the contributions from each plex, a fusion module based on attention mechanism is used for the final prediction. Further details of \method, which are not covered in this section, can be found in Appendix~\ref{architecture_appendix}.

\subsection{Multiplex Graph Representation of RNA 3D Structure}
Each RNA 3D structure consists of atoms that are associated with 3D coordinates and atomic numbers. Based on solely this information, we can define different pairwise interactions between atoms, i.e. according to different cutoff values of the pairwise distances. Then a multiplex graph can be constructed to represent the RNA: For each kind of predefined interaction, we use a plex to contain all atoms as nodes and the interactions as edges. The resulting plexes that share the same group of atoms naturally form a multiplex graph. 

In particular, we are inspired by molecular mechanics~\cite{schlick2010molecular}, in which the molecular energy $E$ is modeled with a separate consideration of local and non-local interactions: $E=E_{\text {local}}+E_{\text {non-local}}$. Motivated by this, we also decouple the modeling of these two kinds of interactions in \method. For local interactions, we can define them by using either chemical bonds or finding the neighbors of each node within a relatively small cutoff distance depending on the given task. For global interactions, we define them by finding the neighbors of each node within a relatively large cutoff distance. As a result, we construct a two-plex multiplex molecular graph $G = \{G_{global}, G_{local}\}$ as shown in Figure~\ref{fig:method}a.

\subsection{Physics-aware Message Passing Modules}
To update the atomic (or node) embeddings, two message passing modules are used in \method, which are \textit{Global Message Passing} and \textit{Local Message Passing} for $G_{global}$ and $G_{local}$ in $G$, respectively. The design of these message passing modules is inspired by physical principles. In molecular mechanics when modeling molecular energy $E=E_{\text {local}}+E_{\text {non-local}}$, the term for local interactions $E_{\text {local}}=E_{\text {bond}}+E_{\text {angle}}+E_{\text {dihedral}}$ includes $E_{\text {bond}}$ that depends on bond lengths, $E_{\text {angle}}$ on bond angles, and $E_{\text {dihedral}}$ on dihedral angles. The term for non-local interactions $E_{\text {non-local}}=E_{\text {electro}}+E_{\text {vdW}}$ includes electrostatic and van der Waals interactions which depend on interatomic distances. So that for the geometric information in molecular mechanics, the local interactions need pairwise distances and angles, while the non-local interactions only need pairwise distances. The message passing modules in \method also use geometric information in this way when modeling these interactions.

As shown in Figure~\ref{fig:method}b, the \textit{Local Message Passing} which updates the node embeddings based on local interactions will incorporate the related adjacency matrix $\textbf{A}_{local}$, pairwise distances $d_{local}$ and angles $\theta_{local}$. The \textit{Global Message Passing} which updates the node embeddings based on global interactions (both non-local and local interactions) will contain only the related adjacency matrix $\textbf{A}_{global}$ and pairwise distances $d_{global}$. The node embeddings learned by each message passing module are passed to the next layer or to the fusion module for final prediction.

\subsection{Fusion Module}
To combine the node embeddings from message passing modules in every hidden layer for final prediction, we design a fusion module as shown in Figure~\ref{fig:method}b. The fusion is a two-step pooling process. In the first step, we use an attention mechanism~\cite{velivckovic2018graph} for each hidden layer $t$ of \method to get the corresponding prediction ${y}_t$ using the node embeddings computed by the two message passing modules in hidden layer $t$. In each attention operation, attention weights for each group of node embeddings are learned and used to compute the corresponding prediction ${y}_t$ using a weighted summation. In the second step, the predictions from all hidden layers are averaged together to compute the final predicted result ${y}$.

\section{Experiments}
\textbf{Task. } In our experiments, we focus on the prediction of 3D RNA structures. Following the previous works~\cite{wang20153drnascore,watkins2020farfar2,townshend2021geometric}, we refer it to the task of identifying accurate structural models of RNA from less accurate ones: Given a group of candidate 3D structural models generated based on an RNA sequence, a desired scoring function can distinguish accurate structural models among all candidates. For PaxNet, only a limited number of RNA structures are used for learning, and no assumptions about structural characteristics are incorporated to perform the identification. In practice, PaxNet predicts the root mean square deviation (RMSD) from the unknown true structure for each structural model. A lower RMSD would suggest a more accurate structural model predicted by PaxNet.

\textbf{Datasets. } To access the performance of PaxNet on predicting 3D RNA structures, we use the same datasets as those used in~\cite{townshend2021geometric}, which include a dataset for training and two benchmarks for evaluation. The \textbf{training dataset} contains 18 relatively older and smaller RNA molecules determined experimentally~\cite{das2007automated}. Each RNA is used to generate 1000 structural models via the Rosetta FARFAR2 sampling method~\cite{watkins2020farfar2}. The two benchmarks for evaluation contain relatively newer and larger RNAs. In detail, \textbf{benchmark 1} consists of the first 21 RNAs in the RNA-Puzzles structure prediction challenge~\cite{miao2020rna}. Each RNA is used to generate at least 1500 structural models using FARFAR2, where 1$\%$ of the models are near native (i.e., within a 2$\angstrom$ RMSD of the experimentally determined native structure). To simulate a difficult modeling scenario, \textbf{benchmark 2} contains no near-native models, and includes 16 RNAs that are all substantially different from any of those used in the aforementioned datasets\footnote{We use the updated version of benchmark 2 provided by the authors of~\cite{townshend2021geometric}, which leads to corrected and different results as those shown in their original paper.}. The source of the datasets can be found in~\cite{townshend2021geometric}. The statistics of the datasets are listed in Appendix~\ref{stats_appendix}.

\begin{figure*}[t]
    \centering
	\includegraphics[width=0.95\columnwidth]{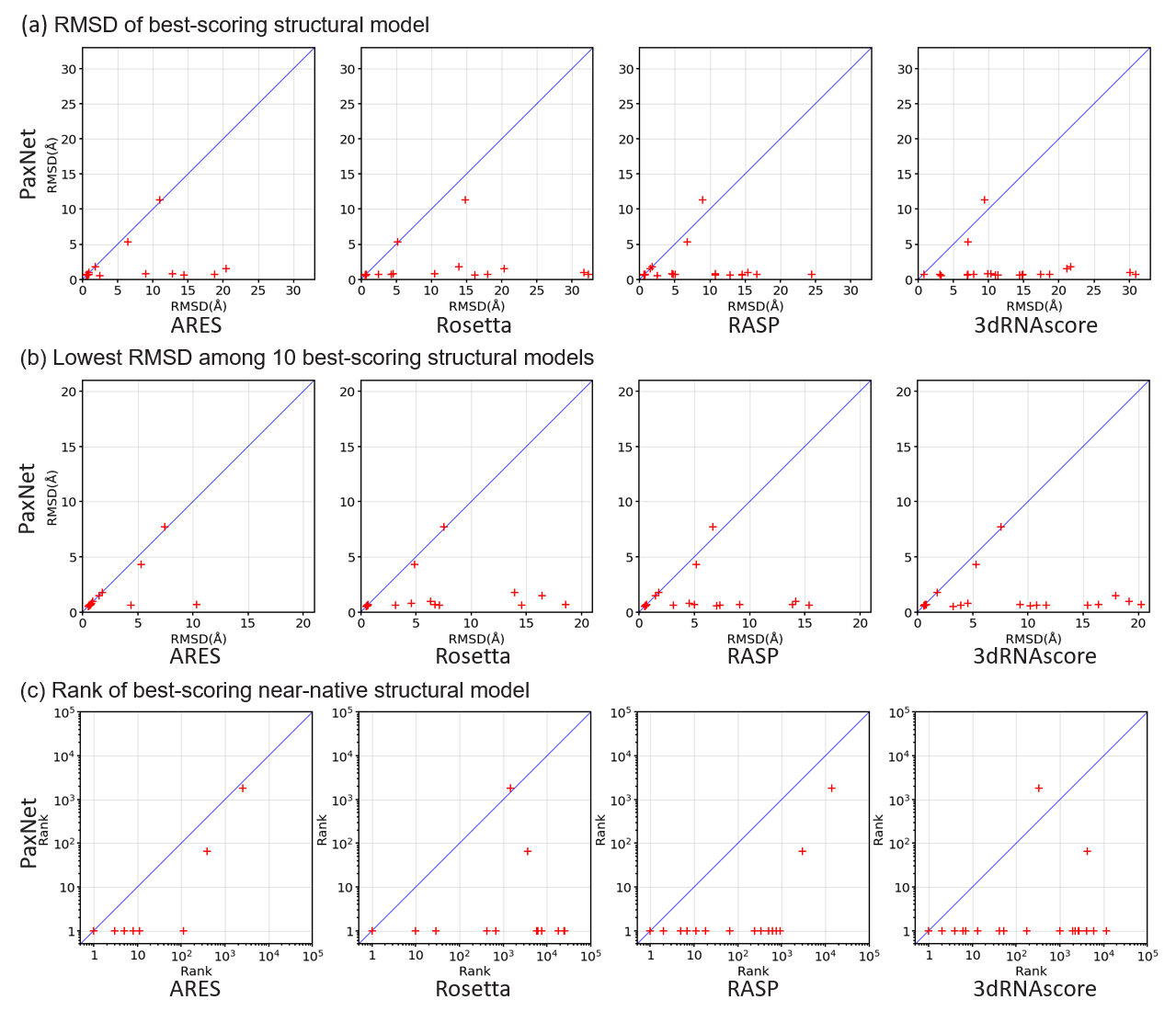}
	\vskip -0.05in
	\caption{\label{fig:benchmark1}\textbf{Performance comparison on benchmark 1.} Given a group of candidate structural models for each RNA in benchmark 1, we rank the models using \method and the other four leading scoring functions for comparison. Each cross in the figures corresponds to one RNA. (a) The best-scoring structural model of each RNA predicted by the scoring functions is compared. \method in general identifies more accurate models (with lower RMSDs from the native structure) than those decided by the other scoring functions. (b) Comparison of the 10 best-scoring structural models. The identifications of \method contain accurate models more frequently than those from other scoring functions. (c) The rank of the best-scoring near-native structural model for each RNA is used for comparison. \method usually performs better than the other scoring functions by having a lower rank.}
	\vskip -0.1in
\end{figure*}

\textbf{Experimental Settings. } We use the same data splits as those used in~\cite{townshend2021geometric} to be the training set and validation set. For the training process, PaxNet is optimized to minimize the difference between the output value and the ground-truth RMSD of each structural model from the corresponding structure on the training set. An early-stopping strategy is adopted to decide the best epoch based on the validation loss. For the benchmark 1, we compare PaxNet with four state-of-the-art baselines: \textbf{ARES}~\cite{townshend2021geometric}, \textbf{Rosetta (2020 version)}~\cite{watkins2020farfar2}, \textbf{RASP}~\cite{capriotti2011all}, and \textbf{3dRNAscore}~\cite{wang20153drnascore}. For benchmark 2, besides the aforementioned baselines, we additionally include \textbf{SimRNA}~\cite{boniecki2016simrna}, \textbf{Rosetta (2010 version)}, and \textbf{Rosetta (2007 version)}. More details of the implementations can be found in Appendix~\ref{implementation_appendix}.

\begin{figure*}[t]
    \centering
	\includegraphics[width=0.75\columnwidth]{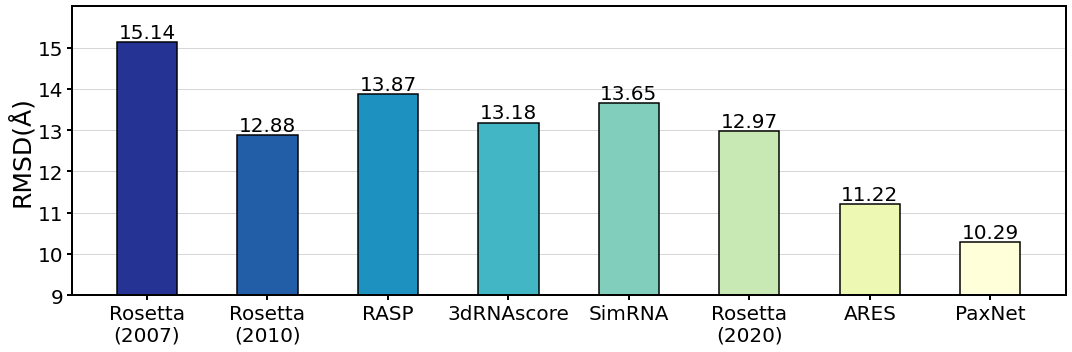}
	\vskip -0.05in
	\caption{\label{fig:benchmark2}\textbf{Performance comparison on benchmark 2.}}
\end{figure*}

\begin{table}[t]
\caption{\textbf{Comparison of space and time complexity.}}
\label{table:complexity}
\centering
\vskip -0.05in
\begin{tabular}{cccc}
	\toprule
	Model & \# of Parameters & Memory (GB) & Inference Time (s)\\
	\midrule
	ARES & 18073 & 13.47 & 2.14\\
	\method & \textbf{12530} & \textbf{7.83} & \textbf{0.57}\\
	\bottomrule
\end{tabular}
\vskip -0.1in
\end{table}

\section{Results}
\textbf{Benchmark 1. }
On benchmark 1, \method significantly outperforms all other four scoring functions as shown in Figure~\ref{fig:benchmark1}. When comparing the best-scoring structural model of each RNA (Figure~\ref{fig:benchmark1}a), the probability of the model to be near-native (<2$\angstrom$ RMSD from the native structure) is 90$\%$ when using \method, compared with 62, 43, 33, and 5$\%$ for ARES, Rosetta, RASP, and 3dRNAscore, respectively. As for the 10 best-scoring structural models of each RNA (Figure~\ref{fig:benchmark1}b), the probability of the models to include at least one near-native model is 90$\%$ when using \method, compared with 81, 48, 48, and 33$\%$ for ARES, Rosetta, RASP, and 3dRNAscore, respectively. When comparing the rank of the best-scoring near-native structural model of each RNA (Figure~\ref{fig:benchmark1}c), the geometric mean of the ranks across all RNAs is 1.7 for \method, compared with 3.6, 73.0, 26.4, and 127.7 for ARES, Rosetta, RASP, and 3dRNAscore, respectively. The lower mean rank of \method indicates that less effort is needed to go down the ranked list of \method to include one near-native structural model. A more detailed analysis of the near-native ranking task can be found in Appendix~\ref{benchmark1_results}. We also conduct an ablation study to demonstrate the effectiveness of the global and local message passing as well as fusion modules in \method. The details of the results are shown in Appendix~\ref{benchmark1_ablation}. 

\textbf{Benchmark 2. }
For each RNA in benchmark 2, we identify the best-scoring structural model scored by each scoring function. For each scoring function, we show the median across RNAs. As shown in Figure~\ref{fig:benchmark2}, \method performs better than all scoring functions by having the lowest median RMSD across RNAs. We find a similar trend when considering the 10 best-scoring structural models for each RNA, which is shown in Appendix~\ref{benchmark2_results}. 

\textbf{Efficiency Evaluation. }
When implementing computational methods for RNAs, a common bottleneck comes from the relatively high complexity caused by the number of involved atoms. Thus we evaluate the efficiency of \method by comparing it with the strongest baseline, ARES. When solving the same given task, we find \method is much more efficient than ARES regarding both space and time complexity as shown in Table~\ref{table:complexity}. Details of the settings can be found in Appendix~\ref{implementation_appendix}. 

\section{Conclusion}
In this work, we focus on the task for RNA 3D structure prediction and propose a GNN-based method, \method, to tackle the task from a deep learning perspective. Different from the existing deep learning-based methods in this field, \method is inspired by the physical principles in molecular mechanics to decouple the modeling of local and non-local interactions in RNA 3D structures. With only atomic types and 3D coordinates and without any assumptions about structural features, \method can be trained on limited RNA 3D structures to make more accurate predictions than the previous state-of-the-art models. The limitation of \method is that we still need an extra sampling method to generate candidate structural models. It would be interesting to extend \method to perform \textit{de novo} RNA 3D structure prediction without a separate sampling process. \method can be also possibly applied to other tasks related to macromolecules due to its general framework.

\newpage

\bibliographystyle{unsrt}
\bibliography{main}

\appendix

\section{Appendix}
\counterwithin{figure}{section}
\counterwithin{table}{section}

\subsection{Notations and Definitions}
\textbf{Notations.}
Let $G = ( V , E )$ be a graph with $N=|V|$ nodes and $M = |E|$ edges. The nearest neighbors of node $i$ are defined as $\mathcal { N } ( i ) = \{ j | d ( i , j ) = 1 \}$, where $d ( i , j )$ is the shortest distance between node $i$ and $j$. In later formulations, we will use $\boldsymbol{h}_{i}$ as the embedding of node $i$, $\boldsymbol{e}_{j i}$ as the edge embedding between node $i$ and $j$, $\boldsymbol{m}_{ji}$ as the message being sent from node $j$ to node $i$ in the message passing scheme, $F$ as the hidden dimension in our model, $\mathrm{MLP}$ as multi-layer perceptron, $\concat$ as concatenation operation, $\odot$ as element wise production and $\boldsymbol{W}$ as weight matrix. 

Here we give the definition of a multiplex molecular graph, which is the format of our RNA representation, as follows:
\begin{definition} \label{def:Multiplex} \textbf{Multiplex Molecular Graph.} 
We denote a molecular structure as an $(L+1)$-tuple $G = (V, E^1, \ldots, E^L)$ where $V$ is the set of nodes (atoms) and for each $l\in\{1, 2, \ldots, L\},$ $E^l$ is the set of edges (molecular interactions) in type $l$ that between pairs of nodes (atoms) in $V$. By defining the graph $G^l = (V,E^l)$ which is also called a plex or a layer, the multiplex molecular graph can be seen as the set of graphs $G = \{G^1, G^2, ..., G^L\}$.
\end{definition}
Next we introduce the message passing scheme~\cite{gilmer2017neural} which is a framework widely used in spatial-based GNNs~\cite{wu2020comprehensive} and is the basis of our \method:
\begin{definition} \label{def:MP} \textbf{Message Passing Scheme.} 
Given a graph $G$, the node feature of each node $i$ is $\boldsymbol{x}_i$, and the edge feature for each node pair $j$ and $i$ is $\boldsymbol{e}_{ji}$. The message passing scheme iteratively updates message $\boldsymbol{m}_{ji}$ and node embedding $\boldsymbol{h}_{i}$ for each node $i$ using the following functions:
\begin{align}
\boldsymbol{m}_{ji}^{t} &= f_{\text {m}}(\boldsymbol{h}_{i}^{t-1}, \boldsymbol{h}_{j}^{t-1}, \boldsymbol{e}_{ji}), \\
\boldsymbol{h}_{i}^{t} &= f_{\text {u}}(\boldsymbol{h}_{i}^{t-1}, \sum\nolimits_{j \in \mathcal{N}(i)} \boldsymbol{m}_{ji}^{t}),
\end{align}
where superscript $t$ denotes the $t$-step iteration, $f_{\text {m}}$ and $f_{\text {u}}$ are learnable functions. For each node $i$, $\boldsymbol{x}_i$ is the input node embedding $\boldsymbol{h}_{i}^{0}$.
\end{definition}

\subsection{Architecture of \method}  \label{architecture_appendix}
\begin{figure*}[t]
    \centering
	\includegraphics[width=1.0\columnwidth]{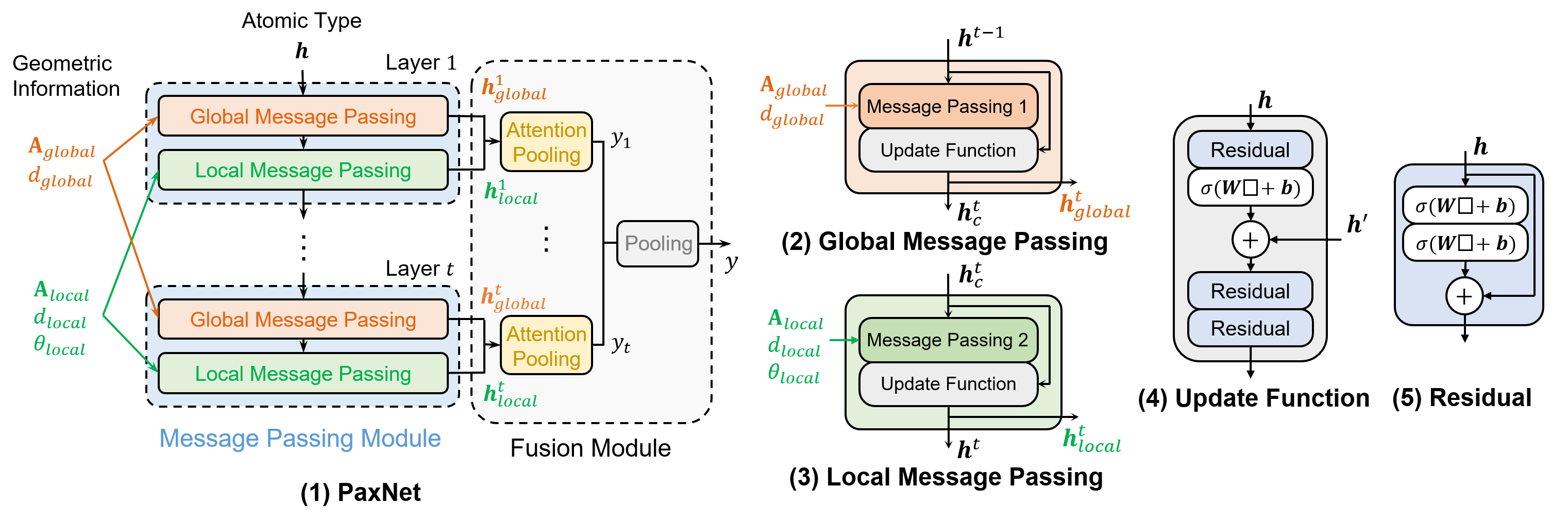}
	\vskip -0.05in
	\caption{\label{fig:paxnet_appendix}\textbf{Illustration of the architecture of \method.} (1) Overall architecture of \method. (2) Architecture of Global Message Passing. (3) Architecture of Local Message Passing. (4) Update function in message passing. (5) Residual block.}
	\vskip -0.1in
\end{figure*}

The architecture of \method is shown in Figure~\ref{fig:paxnet_appendix}. The detailed description of the modules in \method will be covered in this section.

\textbf{Global Message Passing. }
In this module, we update the node embeddings in global plex $G_{global}$ by capturing the pairwise distances $d_{global}$ based on the message passing in Definition~\ref{def:MP}. Each message passing operation is:
\begin{align}
\boldsymbol{m}_{ji}^{t-1} &= \mathrm{MLP}_m([\boldsymbol{h}_{j}^{t-1} \concat \boldsymbol{h}_{i}^{t-1} \concat \boldsymbol{e}_{j i}]),\label{message_embedding}\\
\boldsymbol{h}_{i}^{t} &= \boldsymbol{h}_{i}^{t-1} + \sum\nolimits_{j \in \mathcal{N}(i)} \boldsymbol{m}_{ji}^{t-1}\odot \phi_{d}(\boldsymbol{e}_{j i}), \label{node_update_g}
\end{align} 
where $i, j \in G_{global}$ are connected nodes that define a message embedding, $\phi_{d}$ is a learnable function for pairwise distance. The edge embedding $\boldsymbol{e}_{j i}$ encodes the corresponding pairwise distance information.

After the message passing, an update function containing multiple residual blocks is used to get the node embeddings for the next layer as well as $\boldsymbol {h}_{global}$ to be fed into the fusion module. Illustrations of these operations are shown in Figure~\ref{fig:paxnet_appendix}.

\begin{figure*}[t]
    \begin{center}
	\centerline{\includegraphics[width=1.0\columnwidth]{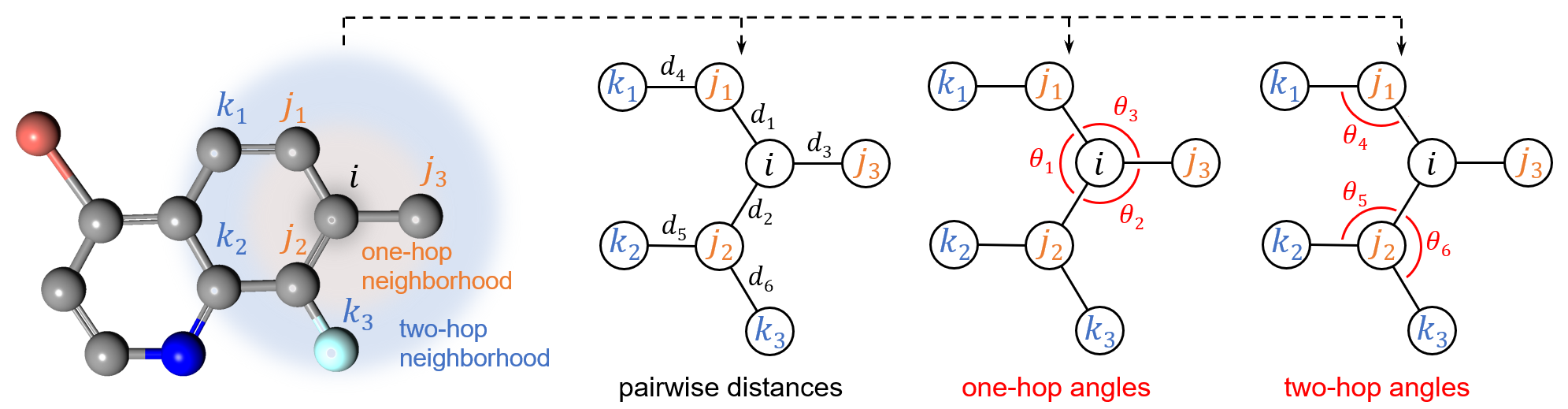}}
	\vskip -0.05in
	\caption{\label{fig:geometric} \textbf{An example of the geometric information in $G$.} By defining the one-hop neighbors $\{j\}$ and two-hop neighbors $\{k\}$ of node $i$, we can define the pairwise distances $d$ and the related angles $\theta$ including one-hop angles and two-hop angles.}
	\end{center}
	\vskip -0.25in
\end{figure*}

\textbf{Local Message Passing. } 
For the updates of node embeddings in the local plex $G_{local}$, we incorporate both pairwise distances $d_{local}$ and angles $\theta_{local}$. When updating the embedding of node $i$, we consider the one-hop neighbors $\{j\}$ and the two-hop neighbors $\{k\}$ of $i$. Specifically for the angles related to those nodes, we show an example in Figure~\ref{fig:geometric} and cluster them depending on the edges around $i$: (a) The \textit {one-hop angles} are angles between the one-hop edges ($\theta_{1}$, $\theta_{2}$ and $\theta_{3}$ in Figure~\ref{fig:geometric}). (b) The \textit {two-hop angles} are angles between the one-hop edges and two-hop edges ($\theta_{4}$, $\theta_{5}$ and $\theta_{6}$ in Figure~\ref{fig:geometric}).

To perform message passing, we use the same way as Equation (\ref{message_embedding}) to compute the message embeddings $\boldsymbol{m}$. The message passing operation in the $t$-th iteration is:
\begin{align}
\boldsymbol{m}_{ji}^{'t-1} = \boldsymbol{m}_{ji}^{t-1} &+ \sum_{j' \in \mathcal{N}(i)\setminus\{j\}} \boldsymbol{m}_{j'i}^{t-1} \odot \phi_{d}(\boldsymbol{e}_{j'i}) \odot \phi_{\theta}(\boldsymbol{\theta}_{j'i, j i}) \nonumber\\
&+ \sum_{k \in \mathcal{N}(j)\setminus\{i\}} \boldsymbol{m}_{kj}^{t-1} \odot \phi_{d}(\boldsymbol{e}_{kj}) \odot \phi_{\theta}(\boldsymbol{\theta}_{k j, j i}), \label{message_update} \\
\boldsymbol{h}_{i}^{t} = \boldsymbol{h}_{i}^{t-1} &+ \sum_{j \in \mathcal{N}(i)} \boldsymbol{m}_{ji}^{'t-1}\odot \phi_{d}(\boldsymbol{e}_{ji}) , \label{node_update_l}
\end{align}
where $i, j, k \in G_{local}$, $\phi_{d}$ is a learnable function for pairwise distance, $\phi_{\alpha}$ is a learnable function for angles. $\boldsymbol{e}_{j i}$ encodes the corresponding pairwise distance information. $\boldsymbol{\theta}_{k j, j i}$ encodes the angle $\theta_{k j, j i}=\angle k j i$ accordingly. In Equation (\ref{message_update}), we use two summation terms to separately encode the one-hop and two-hop angles with the associated pairwise distances to update $\boldsymbol{m}_{j i}$. 

As shown in Figure~\ref{fig:paxnet_appendix}, we design the remaining functions in Local Message Passing similarly to those in Global Message Passing to get $\boldsymbol {h}_{local}$ to be fused and the input for the next iteration.

\textbf{Communication between Message Passings. }
To address the cross-layer relations between $G_{global}$ and $G_{local}$, we let the information in those plexes communicate with each other as depicted in Figure~\ref{fig:paxnet_appendix}. We first perform Global Message Passing on $G_{global}$ in each iteration. Then the updated node embeddings are transferred to $G_{local}$ for Local Message Passing. Finally, the further updated node embeddings are passed back to $G_{global}$ for the next iteration. 

\textbf{Fusion Module. }
As shown in Figure~\ref{fig:paxnet_appendix}, to fuse the node embeddings for a final prediction, we design Fusion Module with a two-step pooling:

In the first step, we use attention mechanism for each hidden layer $t$ of \method to get the corresponding node-level prediction ${y}_t$ using the node embeddings $\boldsymbol {h}_{global}^t$ and $\boldsymbol {h}_{local}^t$ from hidden layer $t$. We first compute the attention weight $\alpha_{m,i}$ that measures the contribution of $\boldsymbol {h}_{m,i}$, which belongs to node $i$ on plex $m$ in $G$:
\begin{align}
\alpha _ {m,i}^ {t} = \frac { \exp (\operatorname { LeakyReLU } (\boldsymbol{W}_m^t\boldsymbol {h}_{m,i}^t)) } { \sum _ { m} \exp ( \operatorname { LeakyReLU } (\boldsymbol{W}_m^t\boldsymbol {h}_{m,i}^t)) },\label{softmax}
\end{align}
where $m \in \{global, local\}$, and $\boldsymbol{W}_m^t \in \mathbb{R}^{1\times F}$ is a learnable weight matrix different for each hidden layer $t$ and plex $m$. With $\alpha _ {m,i}^ {t}$, we then compute the node-level prediction ${y}_{i, t}$ of node $i$ in hidden layer $t$ using weighted summation:
\begin{align}
{y}_{i, t} = \sum\nolimits _ { m } \alpha _ {m,i}^ {t} (\boldsymbol{W}_{out_{m}}^t\boldsymbol {h}_{m,i}^t),\label{weight_sum}
\end{align}
where $\boldsymbol{W}_{out_{m}}^t \in \mathbb{R}^{1\times F}$ is a learnable weight matrix different for each hidden layer $t$ and plex $m$. 

In the second step, the node-level predictions ${y}_{i, t}$ are used to compute the final prediction ${y}$:
\begin{align}
{y} = \frac{1}{NT}\sum\nolimits_{i=1}^{N}\sum\nolimits_{t=1}^{T} {y}_{i, t}.\label{sum}
\end{align} 
where $N$ is the total number of nodes, $T$ is the number of hidden layers in \method.

\begin{table}[t]
\caption{\textbf{Dataset statistics.}}
\label{table:stats}
\centering
\vskip -0.05in
\begin{tabular}{cccc}
	\toprule
	Dataset & \# of RNAs & Avg. \# of Non-hydrogen Atoms & \# of Structural Models\\
	\midrule
	Training Set & 18 & 865 & 18000\\
	Benchmark 1 & 21 & 2172 & 428195\\
	Benchmark 2 & 16 & 1681 & 76442\\
	\bottomrule
\end{tabular}
\vskip -0.05in
\end{table}

\subsection{Experimental Settings}
\subsubsection{Dataset Statistics}   \label{stats_appendix}
The statistics of the datasets used in this paper are summarized in Table~\ref{table:stats}.

\subsubsection{Implementation Details}  \label{implementation_appendix}
When building multiplex graphs for RNA structures, we use cutoff distance $d_l=2.6\angstrom$ for the local interactions in $G_{local}$ and $d_g=20\angstrom$ for the global interactions in $G_{global}$. Following~\cite{townshend2021geometric}, we only use the carbon, nitrogen, and oxygen atoms in RNA structures. For the input features, we use atomic numbers $Z$ as node features and pairwise distances between atoms as edge features computed from atomic positions. We represent $Z$ with randomly initialized, trainable embeddings and expand pairwise distances and angles with basis functions to reduce correlations. For the basis functions $\boldsymbol{e}_{RBF}$ and $\boldsymbol{a}_{SBF}$, we use $N_{\text{SHBF}}$=7, $N_{\text{SRBF}}$=6 and $N_{\text{RBF}}$=16. In our message passing operations, we define $\phi_{d}(\boldsymbol{e})=\boldsymbol{W}_{\boldsymbol{e}}\boldsymbol{e}$ and $\phi_{\alpha}(\boldsymbol{\alpha})=\mathrm{MLP}_{\alpha}(\boldsymbol{\alpha})$, where $\boldsymbol{W}_{\boldsymbol{e}}$ is a weight matrix, $\mathrm{MLP}_{\alpha}$ is a multi-layer perceptrons (MLP). For the MLPs used in our model, they all have 2 layers to take advantage of the approximation capability of MLP. For all activation functions, we use the self-gated Swish activation function. \method is optimized by minimizing the smooth L1 loss between the predicted value and the ground truth using the Adam optimizer. For the best-performed \method, we use a batch size=8, learning rate=1e-4, number of hidden layer=1, number of hidden dimension=16. PyTorch is used to implement \method. All of the experiments are done on an NVIDIA Tesla V100 GPU (32 GB). 

For efficiency evaluation, we use \method and ARES to predict for the structural models of RNA in puzzle 5 of RNA-Puzzles challenge. The RNA being predicted has 6034 non-hydrogen atoms. The model settings of \method and ARES are the same as those used for reproducing the experiments on the benchmarks. We use batch size=8 when performing the predictions.

\subsection{Additional Results}
\subsubsection{Detailed Analysis of Near-native Ranking Task on Benchmark 1}  \label{benchmark1_results}

\begin{figure*}[t]
    \centering
	\includegraphics[width=0.8\columnwidth]{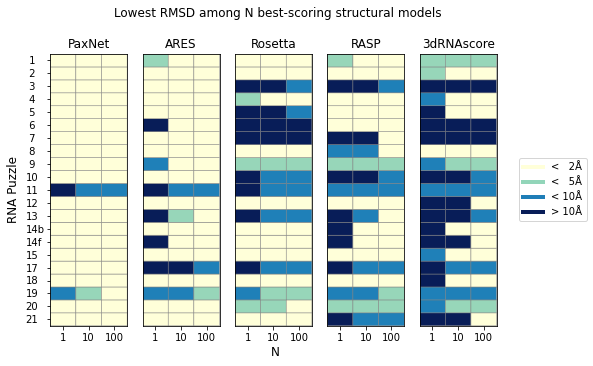}
	\vskip -0.05in
	\caption{\label{fig:benchmark1_appendix}\textbf{Detailed analysis of near-native ranking task on benchmark 1.} The results of the lowest RMSD among N best-scoring structural models of each RNA predicted by each scoring function are compared.}
	\vskip -0.1in
\end{figure*}

For each RNA in benchmark 1, we rank the structural models using \method and four baseline scoring functions. For each scoring function, we select the N $\in \{1, 10, 100\}$ best-scoring structural models for each RNA. For each RNA, scoring function, and N, we show the lowest RMSD across structural models in Figure~\ref{fig:benchmark1_appendix}). The RMSD results are quantized to determine if each RMSD is below 2$\angstrom$, between 2$\angstrom$ and 5$\angstrom$, between 5$\angstrom$ and 10$\angstrom$, or above 10$\angstrom$. From the results, we find that for each RMSD threshold (2$\angstrom$, 5$\angstrom$, or 10$\angstrom$) and for each N, the number of RNAs with at least one selected model that has RMSD below the threshold is greater when using \method than when using any of the other four baseline scoring functions.

\subsubsection{Additional Results on Benchmark 2} \label{benchmark2_results}

\begin{figure*}[t]
    \centering
	\includegraphics[width=0.8\columnwidth]{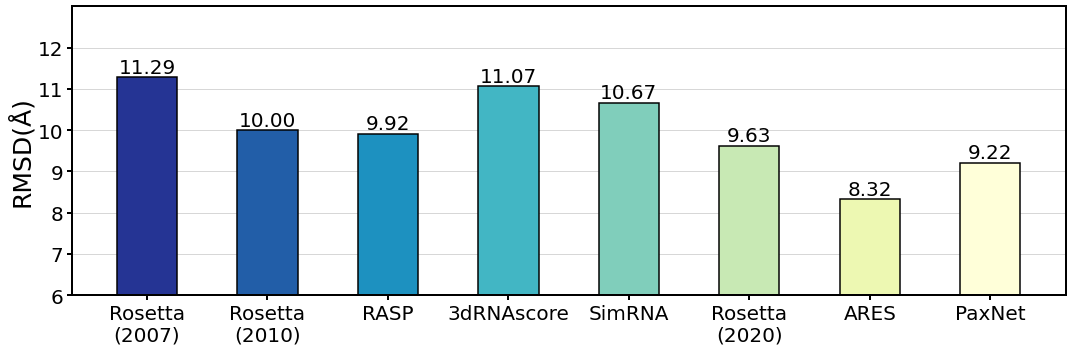}
	\vskip -0.05in
	\caption{\label{fig:benchmark2_appendix}\textbf{Additional Results on Benchmark 2.} For each RNA and for each of the eight scoring functions, we identify the minimum RMSD across the 10 best-scoring structural models. For each scoring function, we show the median across RNAs.}
\end{figure*}

We score all structural models for each RNA and for each of the eight scoring functions, and then identify the minimum RMSD across the 10 best-scoring ones. For each scoring function, we use the median across RNAs for comparison. As shown in Figure~\ref{fig:benchmark2_appendix}, \method outperforms all scoring functions except ARES.

\subsubsection{Ablation Study}  \label{benchmark1_ablation}

\begin{figure*}[t]
    \centering
	\includegraphics[width=0.8\columnwidth]{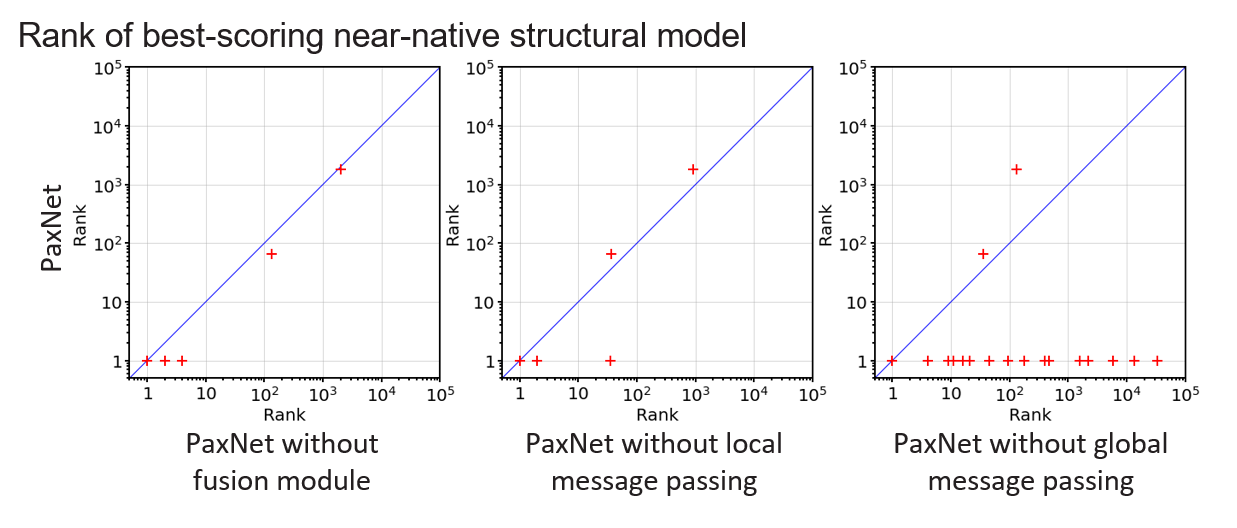}
	\vskip -0.05in
	\caption{\label{fig:ablation_appendix}\textbf{Results of ablation study on benchmark 1.} The rank of the best-scoring near-native structural model for each RNA is used for comparison.}
\end{figure*}

To evaluate the effectiveness of the modules in our \method, we conduct ablation study on benchmark 1 by designing three \method variants: \method without the fusion module (replace the fusion module with average pooling), \method without the local message passing module, and \method without the global message passing module. From the results shown in Figure~\ref{fig:ablation_appendix}, we find the original \method overall performs better than all ablated variants: The geometric mean of the ranks across all RNAs is 1.7 for \method, compared with 2.1, 2.0, and 79.9 for \method without fusion module, \method without local message passing module, and \method without global message passing module, respectively. In particular, \method without global message passing module performs the worst among all ablated variants, which implies the importance of non-local interactions in RNA structures.

\end{document}